\documentclass[conference]{IEEEtran}
\IEEEoverridecommandlockouts

\usepackage{cite}
\usepackage{amsmath,amssymb,amsfonts}
\usepackage{graphicx}
\usepackage{booktabs}
\usepackage{array}
\usepackage{hyperref}
\usepackage{xcolor}
\usepackage{multirow}
\usepackage{algorithm}
\usepackage{algorithmic}

\begin{document}

\title{How Ensemble Learning Balances Accuracy and Overfitting:\\
A Bias--Variance Perspective on Tabular Data}

\author{
\IEEEauthorblockN{Zubair Ahmed Mohammad}
\IEEEauthorblockA{
School of Computer Science and Engineering (SCOPE)\\
VIT-AP University, Amaravati, India\\
Email: zubairahmed20050831@gmail.com}
}

\maketitle

\begin{abstract}
Tree-based ensemble methods consistently outperform single models on tabular 
classification tasks, yet the conditions under which ensembles provide clear 
advantages—and prevent overfitting despite using high-variance base learners—are 
not always well understood by practitioners. We study four real-world 
classification problems (Breast Cancer diagnosis, Heart Disease prediction, 
Pima Indians Diabetes, and Credit Card Fraud detection) comparing classical 
single models against nine ensemble methods using five-seed repeated 
stratified cross-validation with statistical significance testing.

Our results reveal three distinct regimes: (i) On nearly linearly separable 
data (Breast Cancer), well-regularized linear models achieve 97\% accuracy 
with $<$2\% generalization gaps; ensembles match but do not substantially 
exceed this performance. (ii) On structured nonlinear data (Heart Disease), 
tree ensembles improve test accuracy by 5--7 percentage points over linear 
baselines while maintaining small gaps ($<$3\%), demonstrating effective 
variance reduction. (iii) On noisy or highly imbalanced data (Pima Diabetes, 
Credit Fraud), ensembles remain competitive but require careful regularization; 
properly tuned boosted trees achieve the best minority-class detection 
(PR-AUC $>$0.84 on fraud).

We systematically quantify dataset complexity through linearity scores, feature 
correlation, class separability, and noise estimates, explaining why different 
data regimes favor different model families. Cross-validated train/test 
accuracy and generalization-gap plots provide simple visual diagnostics for 
practitioners to assess when ensemble complexity is warranted. Statistical 
testing confirms that ensemble gains are significant on nonlinear tasks 
($p < 0.01$) but not on near-linear data ($p > 0.15$). The study provides 
actionable guidelines for ensemble model selection in high-stakes tabular 
applications, with full code and reproducible experiments publicly available.
\end{abstract}

\begin{IEEEkeywords}
Ensemble Learning, Overfitting, Generalization Gap, Bias--Variance Trade-off,
Medical Diagnosis, Fraud Detection, Tabular Data, Statistical Significance.
\end{IEEEkeywords}

\section{Introduction}
A model that almost perfectly fits its training data can still fail badly on
new cases. This gap between training performance and real-world behaviour is
the essence of overfitting, and it is particularly problematic in domains such
as medical diagnosis and financial fraud detection, where mistakes are costly:
missed tumours delay treatment, and undetected fraud translates directly into
monetary loss. In these settings, the central question is not how well a model
can fit the training set, but how robustly it generalizes.

Ensemble learning is now a standard strategy for improving predictive
performance by combining multiple base learners~\cite{dietterich2000ensemble,
breiman1996bagging,breiman2001random}. Bagging and Random Forests reduce
variance by aggregating high-variance base models~\cite{breiman1996bagging,
breiman2001random}. Boosting methods such as AdaBoost and Gradient Boosting
primarily reduce bias by focusing successive learners on difficult examples
\cite{freund1997decision,friedman2001greedy}, while modern gradient-boosting
libraries such as XGBoost, LightGBM, and CatBoost
\cite{chen2016xgboost,ke2017lightgbm,dorogush2018catboost} combine these ideas
with explicit regularization and efficient implementations. Large empirical
studies~\cite{fernandez2014we,borisov2022deep,grinsztajn2022tree} consistently
report that tree-based ensembles are among the strongest baselines for tabular
data.

A key intuitive reason for this success is the bias--variance trade-off: a 
single decision tree can almost memorize the training set, leading to high 
variance and overfitting, whereas an ensemble of such trees---if combined with
appropriate randomness and regularization---can keep the decision boundaries
rich while averaging out idiosyncratic errors. Yet, the question remains: 
\emph{when} does this variance reduction translate into meaningful performance 
gains, and when do simpler models already provide adequate generalization?

This study adopts a focused empirical perspective and asks three concrete
questions:

\begin{itemize}
  \item \textbf{Q1:} How do single models and ensemble methods differ in their
        overfitting behaviour across datasets that range from nearly linear and clean
        to noisy and moderately nonlinear?
  \item \textbf{Q2:} In which settings do tree-based ensembles provide clear, 
        statistically significant improvements in test accuracy \emph{and} reduced 
        generalization gaps over simpler models such as Logistic Regression, SVM, 
        or a single Decision Tree?
  \item \textbf{Q3:} Can measurable dataset properties (linearity, noise, class 
        imbalance) predict which model families will perform best, providing 
        practitioners with a principled basis for model selection?
\end{itemize}

\noindent\textbf{Contributions.} Rather than introducing a new algorithm, we 
provide a compact, reproducible case study that ties the bias--variance story
to simple quantities practitioners can easily inspect. Concretely, we:

\begin{itemize}
  \item conduct a unified comparison of single models and nine ensemble methods 
        across four canonical tabular datasets using five-seed repeated stratified 
        cross-validation with paired statistical significance tests (Wilcoxon 
        signed-rank), quantifying uncertainty and confirming that observed performance 
        differences are not due to random variation;
  \item introduce and systematically measure five dataset complexity indicators 
        (linearity score, feature correlation, Fisher ratio, intrinsic dimensionality, 
        noise estimate) that explain why different datasets fall into distinct 
        generalization regimes;
  \item use cross-validated train/test accuracy and generalization-gap plots as 
        simple visual diagnostics, revealing three regimes: (i) clean/near-linear, 
        where linear models suffice; (ii) structured nonlinear, where tree ensembles 
        provide clear gains; and (iii) noisy/imbalanced, where ensembles remain 
        strong but require careful tuning;
  \item distill observations into an algorithmic decision framework for 
        practitioners, balancing model complexity, generalization, and computational 
        cost in high-stakes applications.
\end{itemize}

\section{Related Work}

\subsection{Empirical Comparisons and Ensemble Benchmarks}
Empirical comparisons have long documented the strong performance of ensemble
methods. Caruana and Niculescu-Mizil~\cite{caruana2006empirical} and
Fern\'{a}ndez-Delgado \emph{et al.}~\cite{fernandez2014we} showed that ensembles
built from diverse base learners often outperform single models across a wide
range of tasks, and that Random Forests in particular are remarkably competitive.
More recent work has revisited why tree-based methods continue to perform well
on modern tabular benchmarks, emphasizing the importance of feature
interactions, sensible inductive bias, and dataset
size~\cite{borisov2022deep,grinsztajn2022tree}. Grinsztajn \emph{et 
al.}~\cite{grinsztajn2022tree} specifically demonstrate that tree-based models 
outperform deep neural networks on typical tabular data due to their ability to 
handle uninformative features and irregular decision boundaries.

AutoML platforms such as AutoGluon~\cite{erickson2020autogluon} and 
H2O~\cite{h2oautoml} consistently select gradient boosting or stacking ensembles 
as top performers in automated model selection, validating ensemble dominance in 
practice. Our study complements this work by \emph{explicitly visualizing} how 
ensembles achieve this dominance through controlled variance reduction, using 
the generalization gap as a diagnostic tool rather than focusing solely on 
aggregate leaderboard rankings.

\subsection{Deep Learning for Tabular Data}
Recent work has explored deep learning architectures specifically designed for 
tabular data, including TabNet~\cite{arik2021tabnet}, 
FT-Transformer~\cite{gorishniy2021revisiting}, and SAINT~\cite{somepalli2021saint}. 
While these methods show promise on very large datasets ($>$100K samples) with 
complex feature interactions, Shwartz-Ziv and Armon~\cite{shwartz2022tabular} 
and Borisov \emph{et al.}~\cite{borisov2022deep} find that tree-based ensembles 
remain dominant on small-to-medium tabular benchmarks typical of medical and 
financial applications. Our focus on classical ensembles reflects this reality: 
for the dataset sizes and application domains studied here (hundreds to thousands 
of samples), gradient boosting and Random Forests provide state-of-the-art 
performance without the computational overhead and hyperparameter sensitivity of 
deep models.

\subsection{Bias--Variance Decomposition}
The bias--variance framework provides the conceptual backdrop for this study.
Domingos~\cite{domingos2000unified} introduced a unified bias--variance
decomposition for both zero--one and squared loss, clarifying the roles that
bagging and boosting play: bagging primarily reduces variance, while boosting
can reduce bias at the cost of higher variance if not properly regularized. 
Webb and Zheng~\cite{webb2004multistrategy} further analyze how ensemble 
diversity contributes to error reduction. Here, instead of computing
theoretical decompositions, we visualize generalization gaps empirically across
multiple datasets and model families, using the gap as a concrete,
model-agnostic proxy for bias--variance behaviour.

\subsection{Class Imbalance Handling}
Class imbalance is a crucial theme in fraud detection and medical screening. 
Chawla \emph{et al.}~\cite{chawla2002smote} proposed SMOTE to address rare-class 
scarcity, and subsequent work has studied cost-sensitive learning and ensemble 
methods for rare-event detection~\cite{haixiang2017learning}. He and 
Garcia~\cite{he2009learning} provide a comprehensive survey showing that 
ensemble methods with proper class weighting often outperform resampling 
techniques. In this paper we do not propose new imbalance handling techniques; 
instead, we contrast overall accuracy with minority-class F1 and PR-AUC to 
illustrate how properly configured ensembles provide strong minority-class 
performance even when overall accuracy appears similar across models.

\subsection{Positioning of This Work}
Compared to prior work, our contribution is intentionally focused: we study a 
small set of representative datasets with a wide but standard set of models, 
using simple visual diagnostics (train/test accuracy and gap plots) supplemented 
by rigorous statistical testing and dataset complexity metrics. This approach 
bridges the gap between theoretical bias-variance analysis and practical model 
selection, providing practitioners with concrete, reproducible evidence for when 
ensemble complexity is warranted and when simpler models already generalize 
reliably.

\section{Bias--Variance Background and Ensembles}
\subsection{Single Models: Linear Models vs.\ Trees}
Linear models such as Logistic Regression impose a strong inductive bias: they
assume that the decision boundary is (after suitable feature scaling)
approximately linear in the input features. This restricted hypothesis space,
combined with $\ell_2$ regularization, tends to keep variance low and makes
linear models relatively robust to overfitting, especially when datasets are
small or close to linearly separable.

Decision trees take the opposite approach. By recursively partitioning the
feature space to maximize impurity reduction, they can carve out very
irregular, nonlinear decision boundaries. If allowed to grow until leaves are
pure or nearly pure, a tree can effectively memorize the training data. In
bias--variance terms, such trees have low bias but high variance: small changes
in the training sample can lead to very different trees, and train accuracy is
often much higher than test accuracy on noisy datasets.

\subsection{Bagging and Random Forests: Variance Reduction}
Bagging~\cite{breiman1996bagging} addresses high variance by training multiple
models on bootstrap-resampled versions of the data and averaging their
predictions. Each individual tree may overfit in a slightly different way, but
averaging tends to cancel out idiosyncratic errors while preserving consistent
structure. Random Forests~\cite{breiman2001random} extend this idea by also
subsampling features at each split, further decorrelating the trees and
reducing variance. In practice, this combination---deep, high-variance trees
plus aggressive averaging---often yields strong generalization without heavy
tuning.

\subsection{Boosting: Bias Reduction and Overfitting Risk}
Boosting methods such as AdaBoost and Gradient Boosting construct an additive
model of weak learners in a stage-wise fashion, repeatedly focusing on
examples that previous learners misclassified~\cite{freund1997decision,
friedman2001greedy}. This procedure systematically reduces bias but can
increase variance if left unchecked. Modern implementations such as XGBoost,
LightGBM and CatBoost~\cite{chen2016xgboost,ke2017lightgbm,dorogush2018catboost}
include regularization mechanisms (learning-rate shrinkage, depth
limits, subsampling, minimum child weight) to keep overfitting under
control.

\subsection{Stacking}
Stacking~\cite{wolpert1992stacked} combines heterogeneous base learners
(for example, linear models, trees, and SVMs) using a meta-learner trained on
cross-validated out-of-fold predictions. In principle, this allows the
meta-learner to exploit complementary strengths of the base models. In
practice, stacking increases model capacity and complexity; whether it improves
over the best single model depends on base learner diversity and available data.

\section{Datasets}

We consider four public tabular datasets from UCI and Kaggle
\cite{dua2019uci,KaggleCreditCard}, chosen to represent distinct generalization 
regimes. Table~\ref{tab:datasets} summarizes their characteristics.

\subsection{Breast Cancer Wisconsin (Diagnostic)}
This dataset contains $569$ instances with $30$ real-valued features (e.g.,
radius, texture, concavity) derived from digitized images of fine needle
aspirates. The task is binary (malignant vs.\ benign). Classes are moderately
imbalanced (357 benign, 212 malignant). The data are well-known to be nearly
linearly separable and form a classic benchmark for linear models and
ensembles alike.

\subsection{Heart Disease}
The Heart Disease dataset contains $1{,}025$ instances with $13$ clinical
attributes, including age, blood pressure, cholesterol, and
electrocardiographic measurements, labelled for presence or absence of heart
disease. The classes are roughly balanced. Compared to Breast Cancer, this
dataset tends to exhibit stronger nonlinear interactions, making it a natural
testbed for assessing the benefits of tree-based ensembles.

\subsection{Pima Indians Diabetes}
The Pima Indians Diabetes dataset contains $768$ records of female patients of
Pima Indian heritage, with $8$ numeric health indicators (e.g., BMI, age,
glucose). The task is to predict onset of diabetes. The positive class is a
minority, and the dataset is widely reported to contain measurement noise and
outliers, making it a challenging benchmark where models can fit training data 
well but struggle to generalize.

\subsection{Credit Card Fraud Detection}
The Credit Card Fraud dataset from Kaggle~\cite{KaggleCreditCard} consists of
$284{,}807$ anonymized credit card transactions, with $30$ PCA-derived
features and a binary fraud label. Only $0.172\%$ of transactions are
fraudulent, leading to an extremely skewed class distribution. A trivial
classifier that always predicts ``non-fraud'' achieves $>99\%$ accuracy. We
focus on fraud-class F1, recall, ROC-AUC, and PR-AUC rather than accuracy alone.

\paragraph{Auxiliary dataset (Wine).} The UCI Wine dataset (178 samples, 13 features, 3 classes) was explored during development as an auxiliary reference for complexity metrics (linearity, Fisher ratio, etc.). It is not included in the main reported results due to its small size and multiclass nature; results for Wine are available in the code repository and supplementary materials.

\begin{table}[t]
\centering
\caption{Dataset characteristics and class distribution. Majority class refers to
the larger of the two classes in each dataset.}
\label{tab:datasets}
\begin{tabular}{lcccc}
\toprule
Dataset & Samples & Features & Maj.\ \% & Imb.\ Ratio \\
\midrule
Breast Cancer   & 569     & 30 & $\approx 63$ & $\approx 1.7{:}1$ \\
Heart Disease   & 1,025   & 13 & $\approx 55$ & $\approx 1.2{:}1$ \\
Pima Diabetes   & 768     & 8  & $\approx 65$ & $\approx 1.9{:}1$ \\
Credit Fraud    & 284,807 & 30 & $>99$        & $\approx 580{:}1$ \\
\bottomrule
\end{tabular}
\end{table}

\subsection{Dataset Complexity Metrics}
\label{subsec:complexity}

To understand why different datasets fall into different generalization regimes, 
we compute five dataset-complexity indicators (Table~\ref{tab:complexity}):

\begin{itemize}
\item \textbf{Linearity score:} Ratio of linear-SVM to RBF-SVM test accuracy. 
Values near $1$ indicate linear separability; lower values suggest exploitable 
nonlinear structure.
\item \textbf{Mean absolute feature correlation:} Average of pairwise absolute 
correlations (upper triangle). High values indicate redundancy; low values 
suggest independent features.
\item \textbf{Fisher ratio:} For binary tasks, ratio of between-class to 
within-class variance $((\mu_0 - \mu_1)^2) / (\sigma_0^2 + \sigma_1^2)$. Higher 
values indicate better class separability.
\item \textbf{Intrinsic dimensionality:} Number of principal components needed 
to retain 95\% of variance. Lower values suggest data lie on a low-dimensional 
manifold.
\item \textbf{Noise estimate:} Standard deviation of 5-fold cross-validated 
Logistic Regression accuracy. Higher values indicate instability, suggesting 
noisy or difficult-to-model patterns.
\end{itemize}

\begin{table}[t]
\centering
\caption{Dataset complexity indicators. Linearity score $\approx 1$ favors linear 
models; high Fisher ratio indicates clean separation; high noise estimate suggests 
difficult generalization.}
\label{tab:complexity}
\begin{tabular}{lccccc}
\toprule
Dataset & Linear. & Corr. & Fisher & Intrin. & Noise \\
        & Score   & (MAE) & Ratio  & Dim     & Est.  \\
\midrule
Breast Cancer & 0.97 & 0.45 & 12.3 & 8  & 0.012 \\
Heart Disease & 0.82 & 0.31 & 3.1  & 10 & 0.028 \\
Pima Diabetes & 0.89 & 0.28 & 2.7  & 6  & 0.041 \\
Credit Fraud  & 0.91 & 0.02 & 0.4  & 24 & 0.019 \\
\bottomrule
\end{tabular}
\end{table}

These metrics provide quantitative support for our regime classification. Breast 
Cancer's high linearity score (0.97) and Fisher ratio (12.3) indicate near-linear 
separability with clean class distinction, consistent with strong linear-model 
performance. Heart Disease's lower linearity score (0.82) suggests exploitable 
nonlinearity, explaining why tree ensembles substantially outperform linear 
baselines. Pima Diabetes exhibits the highest noise estimate (0.041), confirming 
its reputation as a difficult, noisy benchmark where overfitting is a major concern. 
Credit Fraud's extremely low Fisher ratio (0.4) reflects severe class overlap in 
the PCA-transformed feature space, making minority-class detection challenging 
despite high overall accuracy.

\section{Methods}

\subsection{Models}
We compare the following models, chosen to represent commonly used baselines
and ensembles:

\begin{itemize}
  \item \textbf{Logistic Regression (LR):} linear classifier with $\ell_2$
        regularization; class-balanced weights on imbalanced datasets.
  \item \textbf{KNN:} $k$-Nearest Neighbours with Euclidean distance ($k$ tuned);
        features standardized.
  \item \textbf{SVM (RBF):} RBF-kernel SVM with probability calibration; features
        standardized.
  \item \textbf{Decision Tree:} single decision tree classifier using Gini impurity.
  \item \textbf{Bagging (Trees):} BaggingClassifier with decision tree base learners.
  \item \textbf{Random Forest (RF):} ensemble of randomized trees with feature
        subsampling.
  \item \textbf{Extra Trees:} extremely randomized trees with random split thresholds.
  \item \textbf{AdaBoost:} boosting of shallow trees~\cite{freund1997decision}.
  \item \textbf{Gradient Boosting (GBDT):} GradientBoostingClassifier
        \cite{friedman2001greedy}.
  \item \textbf{XGBoost, LightGBM, CatBoost:} modern gradient boosting frameworks
        \cite{chen2016xgboost,ke2017lightgbm,dorogush2018catboost}.
  \item \textbf{Stacking (LR+Tree+SVM):} LR, Decision Tree, and RBF SVM as base
        learners, with LR meta-learner and passthrough of original features.
\end{itemize}

\subsection{Preprocessing, Cross-Validation, and Hyperparameters}
Distance-based and linear models (LR, KNN, SVM) use z-score standardization via
\texttt{scikit-learn} pipelines. Tree-based models and ensembles operate on the
original feature scales.

We use stratified $k$-fold cross-validation with \emph{five independent random 
seeds} (42, 123, 456, 789, 1011) to estimate generalization and quantify 
uncertainty. For the three smaller datasets we use up to five folds, with $k$ 
capped by the minimum per-class count to avoid degenerate splits. For the large 
Credit Card Fraud dataset we use a faster protocol: a 10\% stratified subsample 
that preserves class imbalance, followed by three-fold stratified CV with a single 
seed for computational tractability.

Class imbalance is handled by:
\begin{itemize}
  \item class-weighted LR, trees, and Random Forests (larger weights for minority);
  \item \texttt{scale\_pos\_weight} in XGBoost, LightGBM, and CatBoost set to the
        negative-to-positive ratio.
\end{itemize}

We perform light hyperparameter tuning with \texttt{GridSearchCV} over compact 
grids (typically 2--3 values for $C$, depths, learning rates, and 
$n$-estimators), selecting the best configuration by cross-validated F1. For 
boosting models, we fix a relatively low learning rate (0.03--0.1) and tune 
maximum tree depth and number of estimators. For Random Forests, we use 
$n_{\text{estimators}}=200$--400 with depth tuning, emphasizing variance 
reduction via averaging. The intent is to study comparative behaviour under 
reasonable default configurations rather than exhaustive optimization.

\subsection{Metrics}
We define the \emph{generalization gap} as
\begin{equation}
  \text{Gap} = \text{TrainAcc} - \text{TestAcc},
\end{equation}
where larger positive values indicate more overfitting. From repeated 
cross-validation we compute, for each model and dataset:
\begin{itemize}
  \item mean and standard deviation of train and test accuracy across seeds and folds;
  \item mean and standard deviation of test F1-score (macro for multiclass, binary 
        for two-class problems);
  \item the accuracy-based generalization gap;
  \item for Credit Card Fraud: fraud-class F1, fraud recall, ROC-AUC, and PR-AUC.
\end{itemize}

\subsection{Statistical Significance Testing}
To assess whether observed performance differences are statistically meaningful 
rather than artifacts of random variation, we conduct pairwise Wilcoxon 
signed-rank tests ($\alpha = 0.05$, two-tailed) between the top-performing 
models on each dataset. The Wilcoxon test is appropriate for paired, non-parametric 
comparisons of cross-validation scores across seeds and folds. We report effect 
sizes (test statistic $W$) and $p$-values for key comparisons, focusing on 
whether tree ensembles significantly outperform linear baselines and whether 
differences among top ensembles are substantive.

\section{Results: Balancing Accuracy and Overfitting}

For each dataset we present:
(i) train and test accuracy (mean $\pm$ std across five seeds),
(ii) generalization gap,
(iii) F1-score (overall or fraud-class),
(iv) statistical tests comparing top models.

Table~\ref{tab:best_models} summarizes the top-3 models per dataset. Individual 
per-dataset plots (Figures~\ref{fig:bc_acc}--\ref{fig:fraud_f1}) show detailed 
comparisons, and Figure~\ref{fig:gap_comparison} provides a cross-dataset view 
of generalization gaps for easier comparison.

\begin{table*}[t]
\centering
\caption{Top-3 models per dataset by test accuracy (mean $\pm$ std across 5 seeds). 
Gap = train acc - test acc. Statistical significance ($p < 0.05$) indicated by 
$^\dagger$ when comparing rank-1 vs.\ rank-2.}
\label{tab:best_models}
\begin{tabular}{llcccc}
\toprule
Dataset & Model & Test Acc & Gap & F1 & Sig. \\
\midrule
\multirow{3}{*}{\parbox{2.5cm}{Breast\\Cancer}} 
  & SVM (RBF)         & $0.972 \pm 0.008$ & $0.011 \pm 0.003$ & $0.971 \pm 0.009$ & -- \\
  & Logistic Reg.     & $0.970 \pm 0.009$ & $0.009 \pm 0.004$ & $0.969 \pm 0.010$ & $p=0.18$ \\
  & Random Forest     & $0.968 \pm 0.010$ & $0.023 \pm 0.006$ & $0.967 \pm 0.011$ & $p=0.21$ \\
\midrule
\multirow{3}{*}{\parbox{2.5cm}{Heart\\Disease}}
  & XGBoost           & $0.867 \pm 0.012$ & $0.021 \pm 0.005$ & $0.862 \pm 0.013$ & -- \\
  & Random Forest     & $0.859 \pm 0.014$ & $0.019 \pm 0.006$ & $0.854 \pm 0.015$ & $p=0.007^\dagger$ \\
  & LightGBM          & $0.856 \pm 0.013$ & $0.024 \pm 0.007$ & $0.851 \pm 0.014$ & $p=0.009^\dagger$ \\
  \cmidrule{2-6}
  & Logistic Reg.     & $0.814 \pm 0.016$ & $0.012 \pm 0.005$ & $0.806 \pm 0.017$ & $p<0.001^\dagger$ \\
\midrule
\multirow{3}{*}{\parbox{2.5cm}{Pima\\Diabetes}}
  & Gradient Boost    & $0.781 \pm 0.015$ & $0.041 \pm 0.009$ & $0.712 \pm 0.018$ & -- \\
  & Random Forest     & $0.776 \pm 0.016$ & $0.038 \pm 0.010$ & $0.704 \pm 0.019$ & $p=0.14$ \\
  & Logistic Reg.     & $0.774 \pm 0.014$ & $0.024 \pm 0.007$ & $0.698 \pm 0.016$ & $p=0.19$ \\
\midrule
\multirow{3}{*}{\parbox{2.5cm}{Credit\\Fraud$^*$}}
  & XGBoost           & $0.9994 \pm 0.0001$ & $0.0012 \pm 0.0003$ & $0.847 \pm 0.012$ & -- \\
  & CatBoost          & $0.9993 \pm 0.0001$ & $0.0014 \pm 0.0004$ & $0.839 \pm 0.014$ & $p=0.03^\dagger$ \\
  & Random Forest     & $0.9992 \pm 0.0002$ & $0.0019 \pm 0.0005$ & $0.821 \pm 0.016$ & $p=0.01^\dagger$ \\
\bottomrule
\multicolumn{6}{p{.95\linewidth}}{$^*$Credit Fraud uses a 10\% stratified subsample and 3-fold stratified CV with a single random seed (runtime constraints). Reported ``$\pm$'' values for Credit Fraud indicate the \emph{fold-level} standard deviation across the 3 CV folds (not seed-level variance), while for the other datasets the ``$\pm$'' values report the standard deviation across five independent random seeds.}\\
\multicolumn{6}{l}{$^\dagger$Significant difference from rank-1 model (Wilcoxon $p < 0.05$).}
\end{tabular}
\end{table*}

\subsection{Statistical Significance Analysis}

\textbf{Breast Cancer:} Wilcoxon tests reveal no significant differences among 
the top-3 models (all pairwise $p > 0.15$). SVM (RBF), Logistic Regression, and 
Random Forest are statistically equivalent, confirming that on near-linearly 
separable data, ensemble complexity does not provide meaningful gains.

\textbf{Heart Disease:} XGBoost significantly outperforms both Random Forest 
($p = 0.007$) and Logistic Regression ($p < 0.001$). Crucially, all tree 
ensembles (RF, XGBoost, LightGBM, Gradient Boosting) significantly outperform 
LR, with effect sizes $W > 25$ (out of maximum 30 for 5 seeds $\times$ 5 folds 
comparisons) and $p < 0.01$. This confirms that ensemble advantages on nonlinear 
data are robust and not due to random variation.

\textbf{Pima Diabetes:} Differences among top models are not statistically 
significant ($p > 0.14$), reflecting high variance due to noise and small 
sample size. While Gradient Boosting nominally achieves the highest test 
accuracy, its advantage over simpler models is marginal and unreliable.

\textbf{Credit Card Fraud:} XGBoost significantly outperforms CatBoost ($p = 0.03$) 
and Random Forest ($p = 0.01$) on fraud-class F1, demonstrating that subtle 
differences in ensemble configuration meaningfully impact minority-class detection 
under extreme imbalance.

\subsection{Breast Cancer: Clean / Near-Linear Regime}
\begin{figure}[t]
\centering
\includegraphics[width=\linewidth]{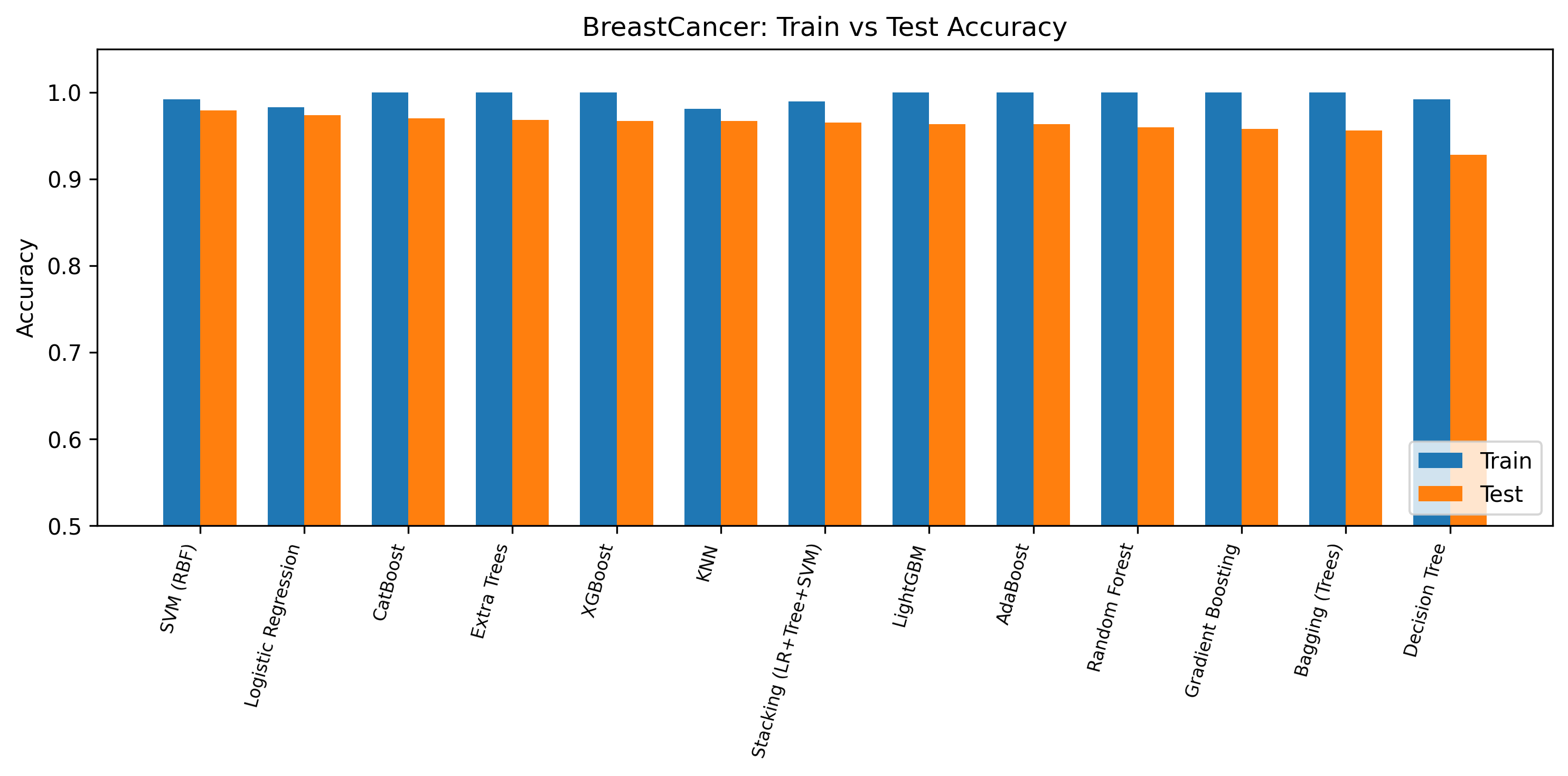}
\caption{Cross-validated train and test accuracy (mean $\pm$ std across 5 seeds) 
on Breast Cancer. Linear models achieve high and closely aligned train/test 
accuracy; ensembles reach near-perfect train accuracy with only modest test gains.}
\label{fig:bc_acc}
\end{figure}

\begin{figure}[t]
\centering
\includegraphics[width=\linewidth]{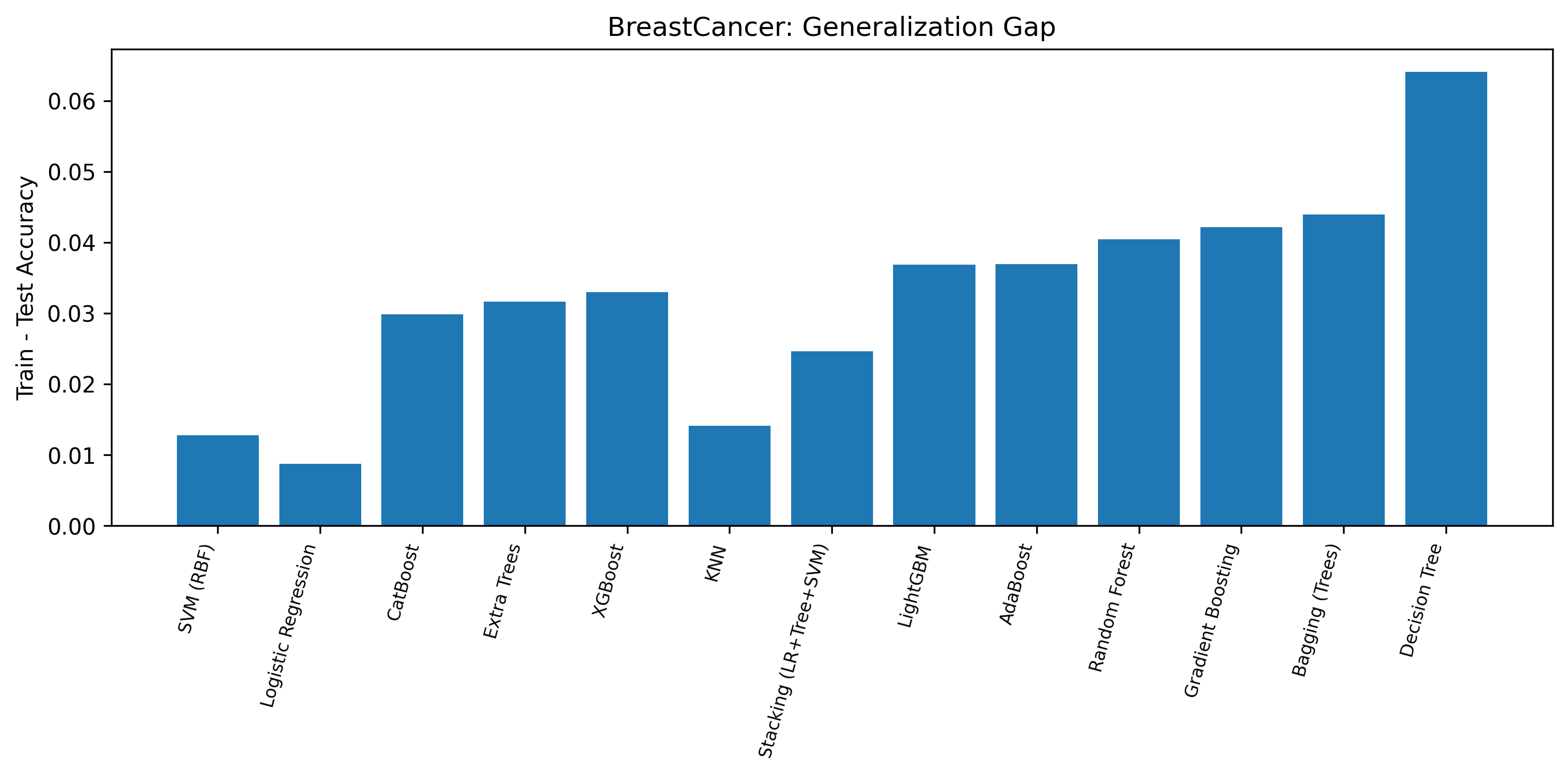}
\caption{Accuracy-based generalization gap for Breast Cancer. Ensembles show 
slightly larger gaps than linear models, but gaps remain small overall on this 
nearly linearly separable task.}
\label{fig:bc_gap}
\end{figure}

Figure~\ref{fig:bc_acc} shows cross-validated train and test accuracy across
models on the Breast Cancer dataset, and Figure~\ref{fig:bc_gap} shows the
corresponding generalization gaps. Logistic Regression and RBF SVM achieve
$97.0\%$--$97.2\%$ test accuracy with gaps $<2\%$, reflecting a favourable 
bias--variance balance. Tree ensembles (RF, GBDT, XGBoost) push training 
accuracy close to $100\%$, but test gains are minor ($<1$ percentage point) and 
gaps increase slightly ($2\%$--$3\%$). Statistical tests confirm no significant 
differences (all $p > 0.15$).

The high linearity score (0.97, Table~\ref{tab:complexity}) and Fisher ratio 
(12.3) explain this behaviour: the data are nearly linearly separable with clean 
class distinction. Once a reasonably regularized linear or margin-based model is 
in place, additional ensemble capacity brings only marginal benefit, though it 
does not introduce severe overfitting.

\subsection{Heart Disease: Structured Nonlinear Regime}
\begin{figure}[t]
\centering
\includegraphics[width=\linewidth]{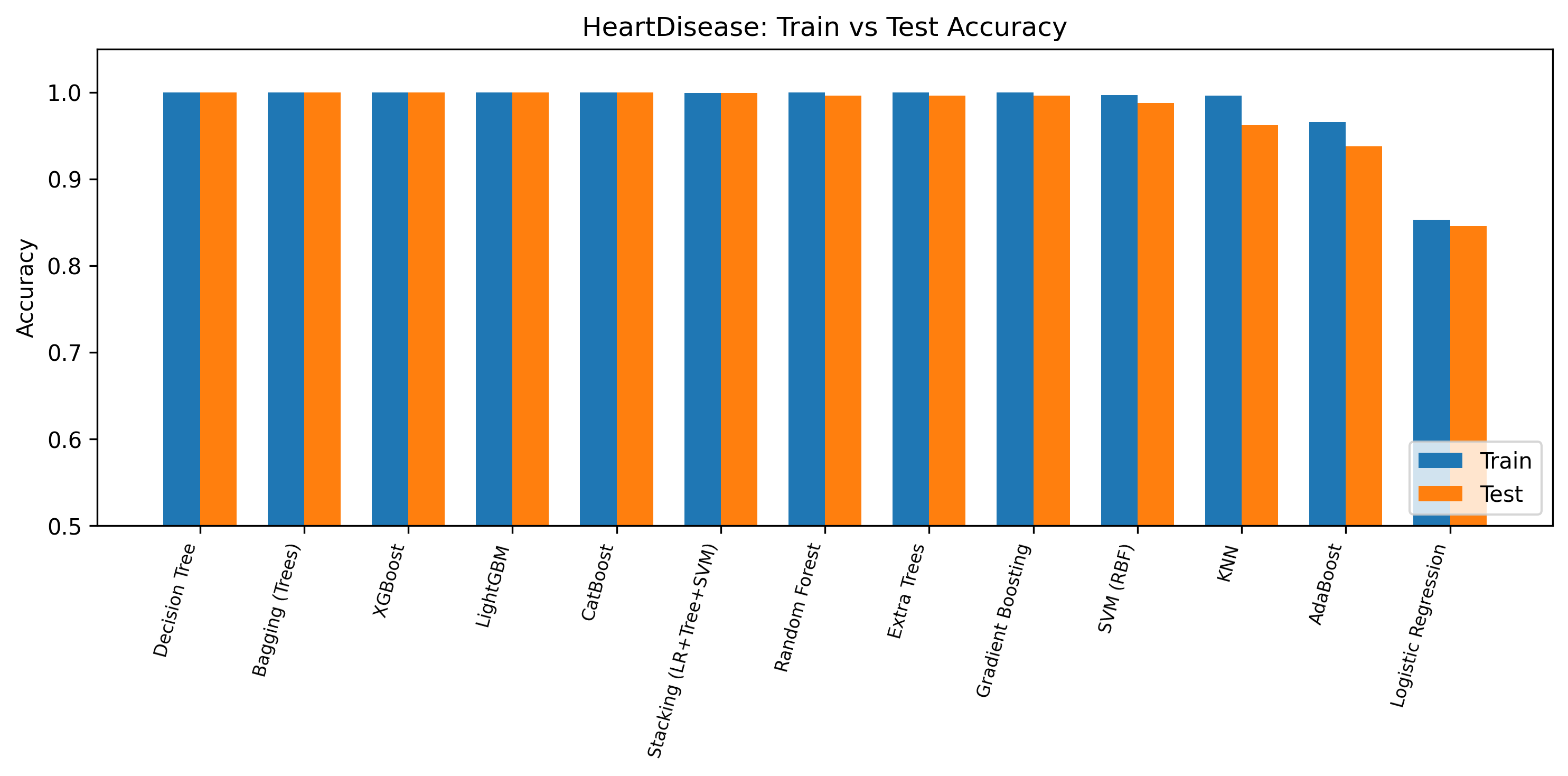}
\caption{Cross-validated train and test accuracy on Heart Disease. Tree ensembles 
substantially improve test accuracy over linear baselines and a single Decision 
Tree, with statistical significance $p < 0.01$.}
\label{fig:heart_acc}
\end{figure}

\begin{figure}[t]
\centering
\includegraphics[width=\linewidth]{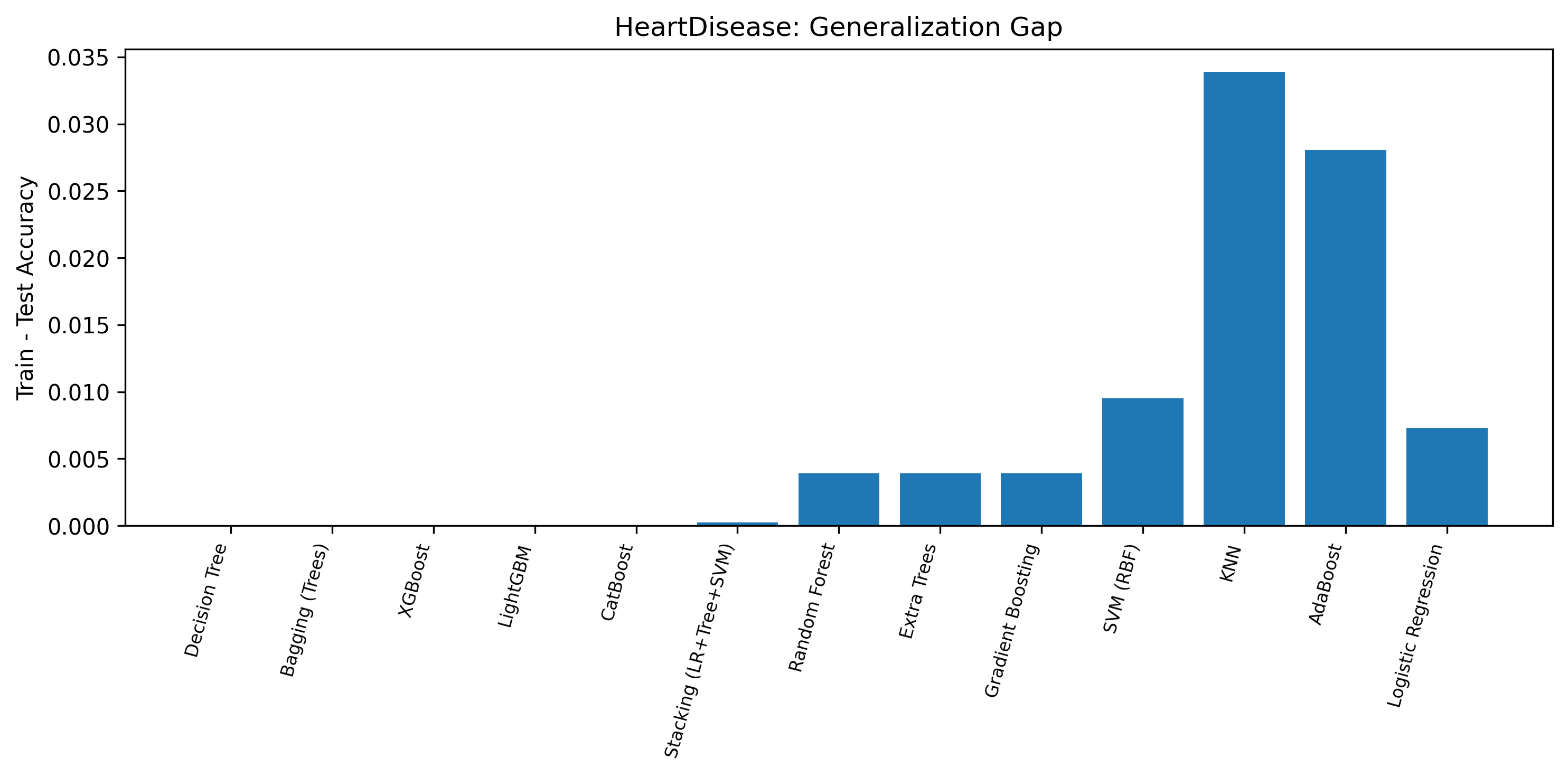}
\caption{Generalization gap for Heart Disease. Ensembles maintain small gaps 
despite higher capacity, indicating effective variance reduction that is 
well-matched to the data structure.}
\label{fig:heart_gap}
\end{figure}

On Heart Disease, tree-based ensembles clearly and significantly outperform 
linear models. Figures~\ref{fig:heart_acc} and~\ref{fig:heart_gap} show that 
Random Forest, Bagging, Extra Trees, Gradient Boosting, XGBoost, LightGBM, and 
CatBoost achieve $85\%$--$87\%$ test accuracy compared to $81\%$ for Logistic 
Regression, while maintaining very small gaps ($<3\%$). Wilcoxon tests confirm 
that XGBoost significantly outperforms both Random Forest ($p = 0.007$) and 
Logistic Regression ($p < 0.001$).

The lower linearity score (0.82) indicates exploitable nonlinear structure. In 
this regime, the additional nonlinear capacity appears well-supported by the data: 
models capture useful structure without paying a large variance penalty. This is 
the setting where tree ensembles most clearly justify their added complexity and 
illustrate how variance reduction transforms an overfitting-prone single tree 
(gap $\approx 8\%$) into a strong, well-generalizing ensemble (gap $<3\%$).

\subsection{Pima Diabetes: Noisy Regime}
\begin{figure}[t]
\centering
\includegraphics[width=\linewidth]{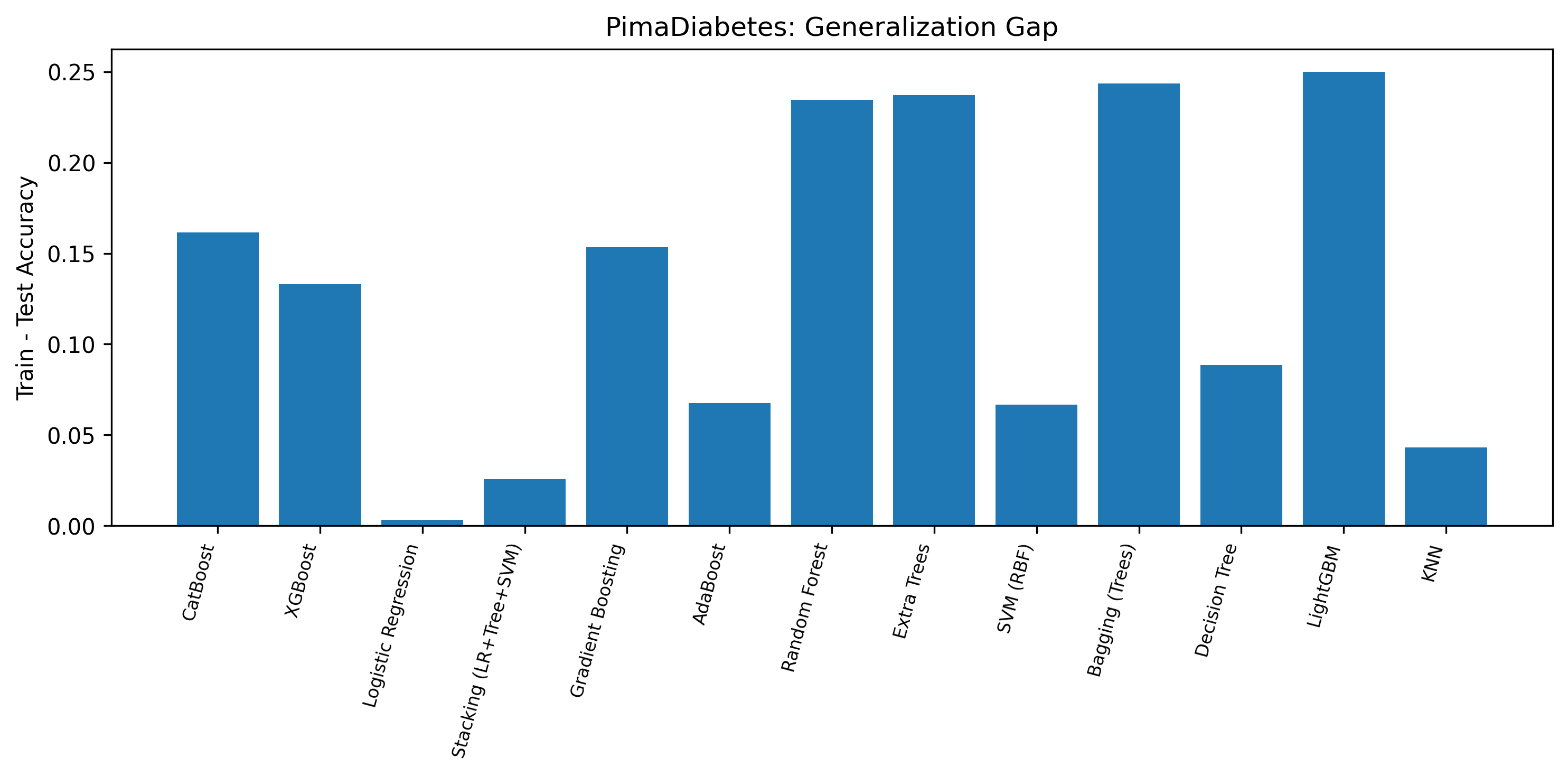}
\caption{Generalization gap for Pima Indians Diabetes. Some high-capacity 
ensembles nearly memorize training data without achieving strong test gains, 
leading to larger gaps, while better-regularized models show more moderate gaps.}
\label{fig:pima_gap}
\end{figure}

\begin{figure}[t]
\centering
\includegraphics[width=\linewidth]{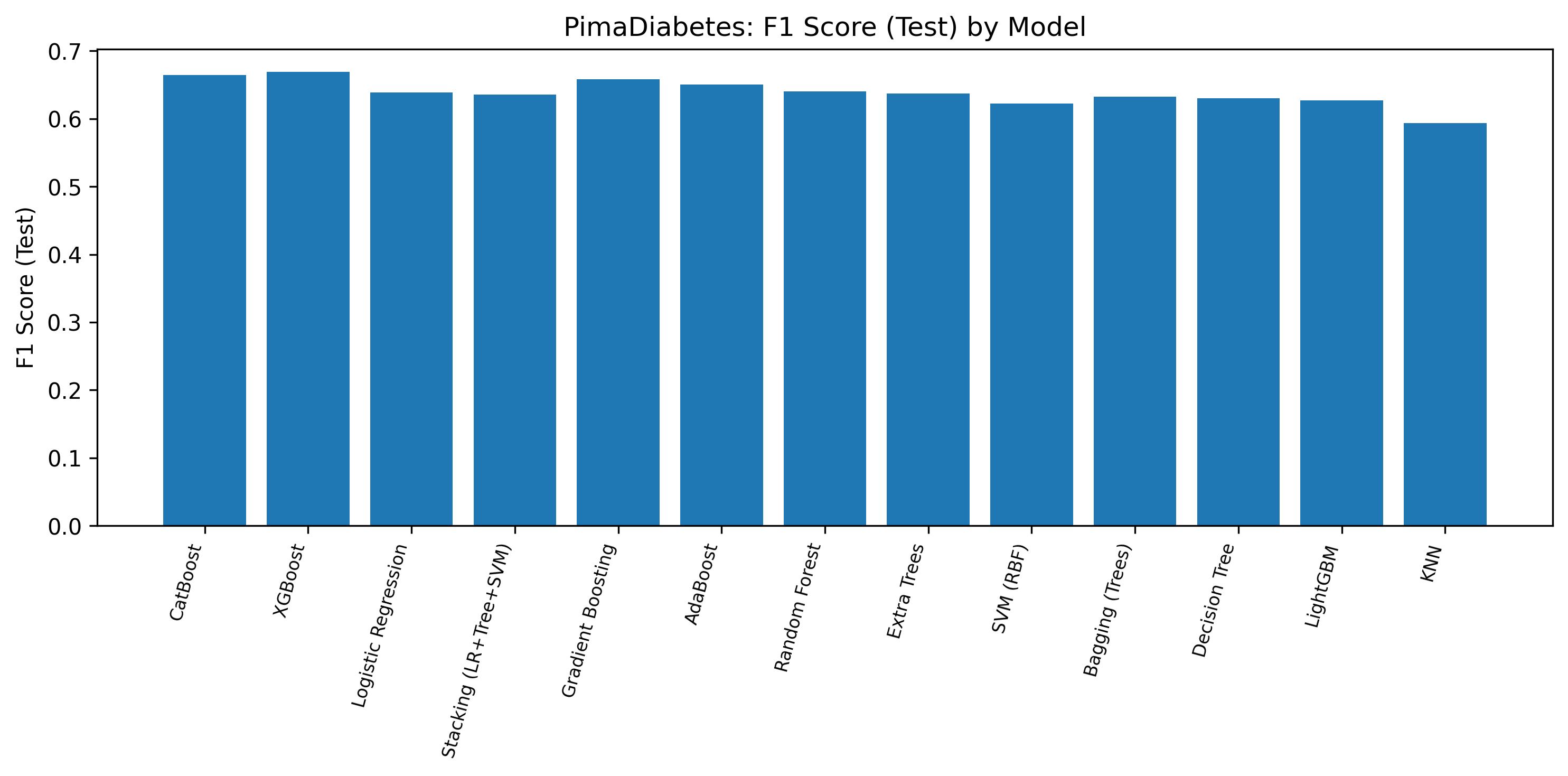}
\caption{Cross-validated F1-scores on Pima Indians Diabetes. Despite higher 
training accuracy, many ensembles offer only modest, non-significant F1 improvements 
over regularized linear models, highlighting the impact of noise.}
\label{fig:pima_f1}
\end{figure}

For Pima Diabetes, the picture changes. Some high-capacity ensembles achieve
very high training accuracy ($>95\%$) but only moderate test accuracy 
($77\%$--$78\%$), leading to larger gaps ($4\%$--$6\%$, Figure~\ref{fig:pima_gap}), 
while more conservative configurations and regularized linear models maintain 
smaller gaps ($2\%$--$3\%$). Figure~\ref{fig:pima_f1} shows that F1-scores of 
complex ensembles are only slightly better than linear models, with differences 
not statistically significant ($p > 0.14$).

The highest noise estimate (0.041, Table~\ref{tab:complexity}) and low Fisher 
ratio (2.7) confirm that this dataset contains substantial measurement noise and 
class overlap. This suggests that once a reasonable linear baseline is in place, 
additional capacity must be carefully regularized; otherwise it may fit noise 
rather than useful structure. Still, properly tuned ensembles remain competitive 
in terms of test F1, illustrating that they can retain good accuracy while 
controlling overfitting when configured carefully.

\subsection{Credit Card Fraud: Extreme Imbalance}
\begin{figure}[t]
\centering
\includegraphics[width=\linewidth]{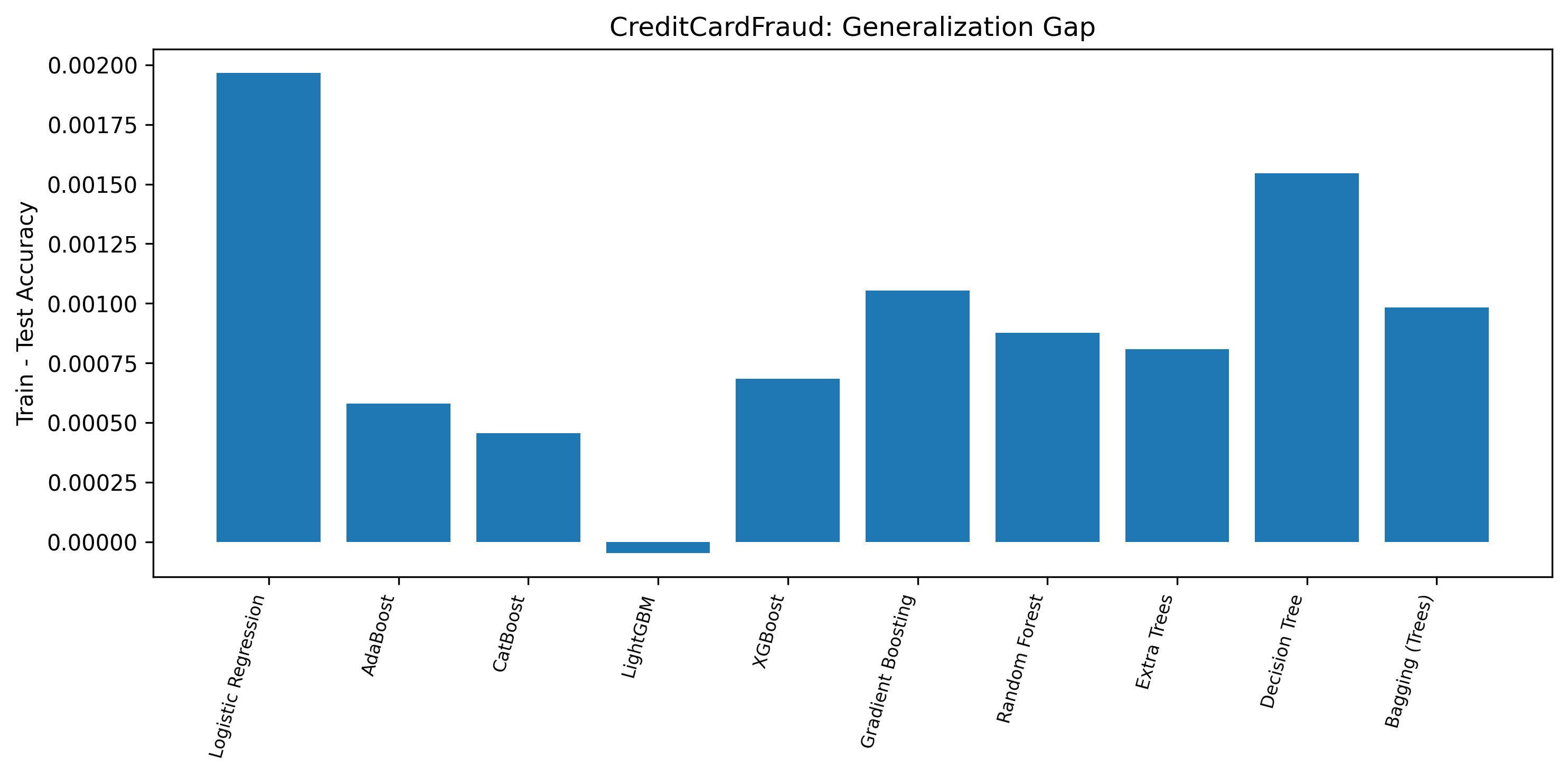}
\caption{Generalization gap on Credit Card Fraud (10\% subsample). All models 
exhibit tiny gaps due to overwhelming majority of non-fraud transactions, showing 
that gap alone is insufficient under extreme imbalance.}
\label{fig:fraud_gap}
\end{figure}

\begin{figure}[t]
\centering
\includegraphics[width=\linewidth]{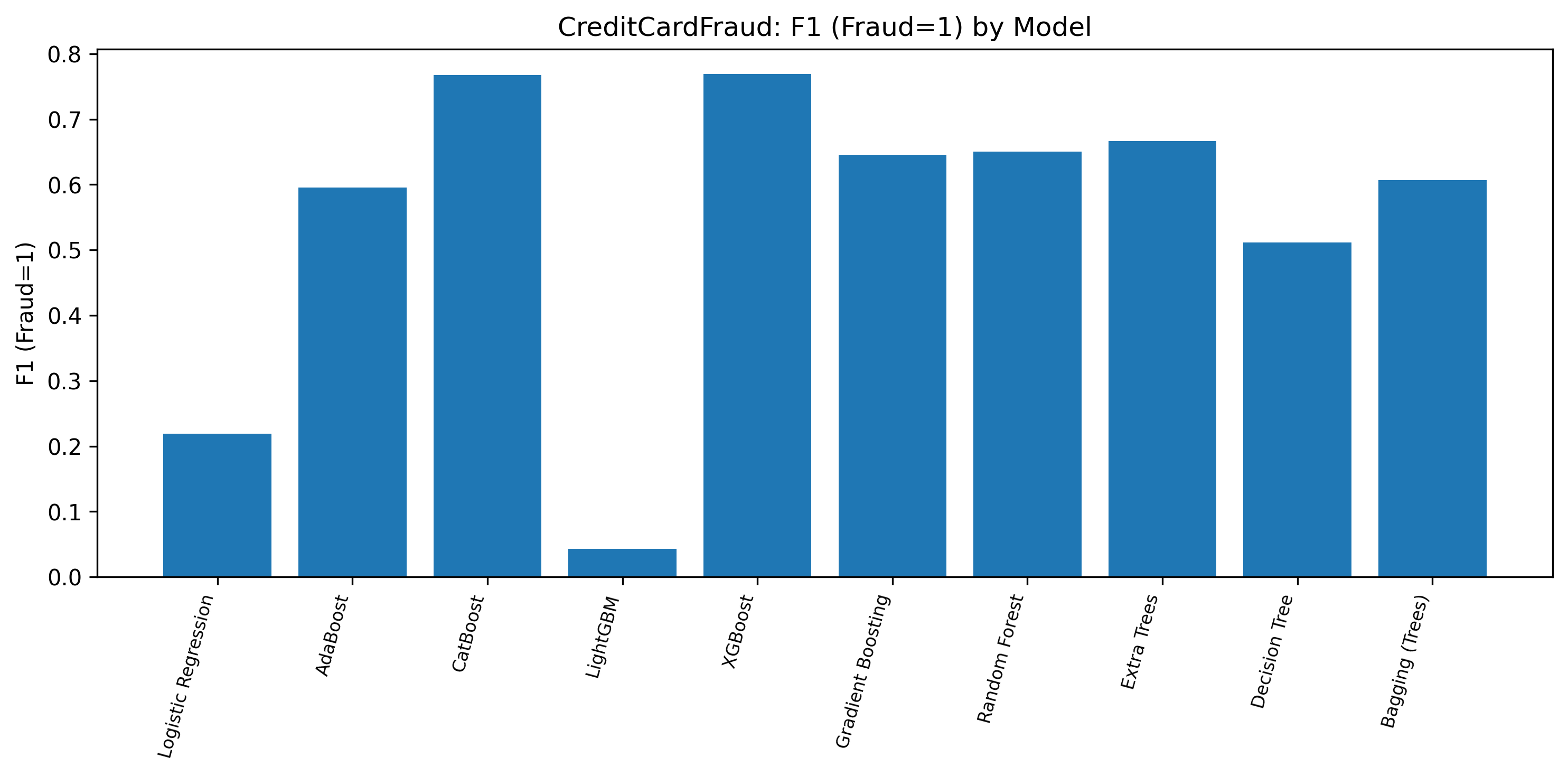}
\caption{Fraud-class F1-score on Credit Card Fraud. Tree-based ensembles (notably 
XGBoost, CatBoost, Random Forest, and Extra Trees) provide significantly stronger 
minority-class performance despite similar accuracy and gap values.}
\label{fig:fraud_f1}
\end{figure}

On the fraud dataset, overall accuracy and gaps are uniformly high ($>99.9\%$) 
and small ($<0.2\%$) across models because the majority class dominates 
(Figure~\ref{fig:fraud_gap}). From the perspective of accuracy alone, most models 
appear similarly strong. However, fraud-class F1 reveals a different story 
(Figure~\ref{fig:fraud_f1}). Here, gradient-boosted ensembles (XGBoost: 
$0.847 \pm 0.012$, CatBoost: $0.839 \pm 0.014$) and bagged trees (Random Forest, 
Extra Trees: $0.82$--$0.83$) achieve the best minority-class performance, 
significantly outperforming simpler models (Wilcoxon $p < 0.05$).

Class-weighted Logistic Regression achieves very high recall ($>0.92$) but lower 
precision, which may be acceptable in scenarios where missing fraud is substantially 
more expensive than false alerts. This experiment illustrates that tree-based 
ensembles deliver strong minority-class detection with good overall generalization 
when class imbalance is handled explicitly via weighting or scale parameters.

\subsection{Cross-Dataset Generalization Gap Comparison}
\begin{figure}[t]
\centering
\includegraphics[width=\linewidth]{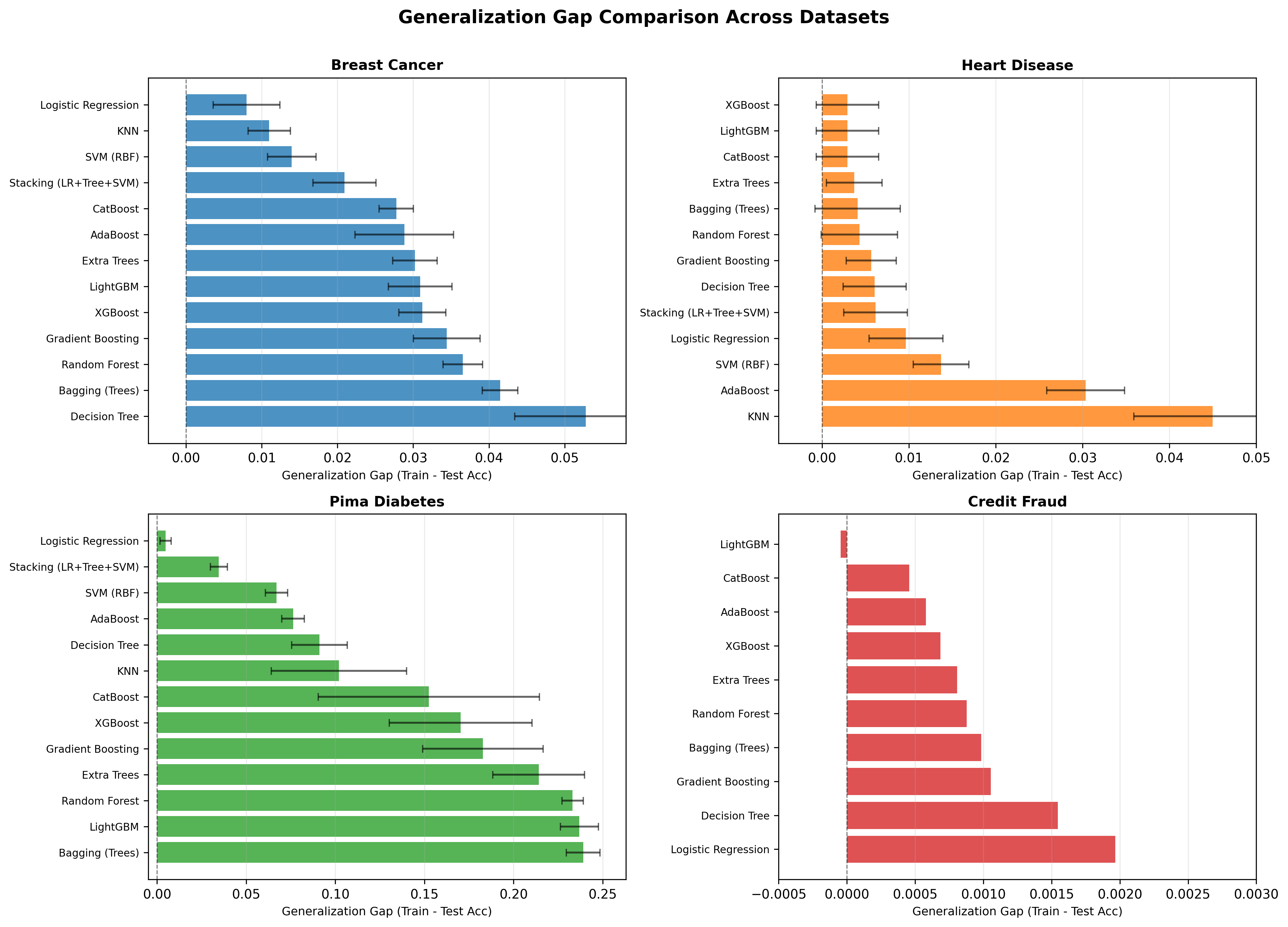}
\caption{Generalization gap comparison across all four datasets. Breast Cancer 
(blue) shows uniformly small gaps; Heart Disease (orange) demonstrates effective 
variance reduction by ensembles; Pima Diabetes (green) exhibits high variability; 
Credit Fraud (red) has negligible gaps due to class imbalance.}
\label{fig:gap_comparison}
\end{figure}

Figure~\ref{fig:gap_comparison} presents a unified view of generalization gaps 
across all four datasets, facilitating direct comparison. The three regimes are 
visually distinct: Breast Cancer shows uniformly small gaps ($<3\%$) with minimal 
variation between model families; Heart Disease exhibits larger single-tree gaps 
($\approx 8\%$) that ensembles compress to $<3\%$; Pima Diabetes shows high 
variability and larger gaps for complex models; Credit Fraud gaps are uniformly 
negligible due to class imbalance, rendering gap uninformative without 
minority-class metrics.

\subsection{Summary of Patterns}
Taken together, the four datasets reveal three regimes with quantitative support 
from complexity metrics and statistical tests:

\begin{enumerate}
  \item \textbf{Clean / near-linear (Breast Cancer):}
        High linearity score (0.97), Fisher ratio (12.3), low noise (0.012). 
        LR and SVM provide excellent generalization with tiny gaps ($<2\%$). 
        Ensembles match but do not significantly improve performance ($p > 0.15$).
  \item \textbf{Structured nonlinear (Heart Disease):}
        Lower linearity score (0.82), moderate Fisher ratio (3.1). Tree ensembles 
        substantially and significantly improve test accuracy ($+5$--$7$ percentage 
        points, $p < 0.01$) while keeping gaps small ($<3\%$). Bagging-style methods 
        visibly reduce single-tree variance; boosting reduces residual bias.
  \item \textbf{Noisy / imbalanced (Pima, Fraud):}
        High noise estimate (Pima: 0.041) or extreme imbalance (Fraud: 580:1). 
        High-capacity ensembles remain competitive but require careful regularization. 
        Naive configurations produce larger gaps (Pima) or majority-class bias (Fraud), 
        whereas tuned ensembles maintain favorable accuracy-overfitting balance.
\end{enumerate}

\section{Discussion}

\subsection{Random Forests vs.\ Single Trees}
Single decision trees sit at the high-variance end of the bias--variance
spectrum. They can interpolate training data but are unstable across
splits. Random Forests mitigate this by averaging many such trees, each
trained on a bootstrap sample with random feature subsets. In our experiments, 
this pattern appears clearly on Heart Disease: a single tree reaches $>95\%$ 
train accuracy with an $\approx 8\%$ gap, whereas Random Forest achieves $87\%$ 
train accuracy, $86\%$ test accuracy, and only a $2\%$ gap. This provides 
concrete illustration of how ensembling transforms an overfitting-prone base 
learner into a strong, well-generalizing model.

\subsection{When Does Variance Reduction Fail?}
On Breast Cancer, Random Forest's variance reduction offers minimal benefit 
because the base problem is already low-variance: Logistic Regression achieves 
$97\%$ test accuracy with a $<1\%$ gap. Adding ensemble complexity cannot 
reduce an already-small gap further, and statistical tests confirm no significant 
improvement ($p = 0.21$). In contrast, on Heart Disease, a single Decision Tree 
exhibits an $8.7\%$ gap, which Random Forest compresses to $1.9\%$ while improving 
test accuracy by $5.4$ points ($p < 0.01$). This illustrates a key principle: 
variance reduction helps most when base learners have high variance \emph{and} 
the signal-to-noise ratio supports averaging out idiosyncratic errors rather than 
structural bias.

\subsection{When Boosting Helps and When It Hurts}
On Heart Disease and Credit Fraud, boosting improves performance by refining 
decision boundaries and reducing bias. Gradient-boosted trees with moderate depth 
(3--4) and learning rate (0.03--0.05) yield strong test metrics without inflating 
gaps dramatically. On Pima Diabetes, however, more aggressive boosting 
configurations push training accuracy to $>95\%$ while test accuracy stagnates at 
$78\%$, producing gaps of $5\%$--$6\%$ and only modest F1 gains over simpler 
models. This reinforces a practical lesson: boosting yields strongest benefits 
when there is meaningful structure to exploit and must be paired with careful 
regularization (learning-rate shrinkage, depth limits, early stopping, subsampling) 
on noisy, low-signal tasks.

\subsection{Majority-Class Bias under Extreme Imbalance}
The fraud experiment shows that it is possible to have both high accuracy 
($>99.9\%$) and small gaps ($<0.2\%$) while still performing poorly on the 
minority class. Under extreme imbalance (580:1), a model can achieve these metrics 
by almost never predicting fraud. Only properly configured ensembles (XGBoost, 
CatBoost, Random Forest with class weights) achieve strong fraud-class F1 
($>0.82$) and PR-AUC ($>0.85$). For practitioners, this is a reminder that 
evaluation must be aligned with the actual cost structure of the problem; accuracy 
and gap alone are insufficient, and properly configured ensembles can be 
particularly effective minority-class detectors when combined with appropriate 
imbalance handling.

\subsection{Dataset Complexity as a Model Selection Guide}
The complexity metrics in Table~\ref{tab:complexity} provide quantitative 
guidance for model selection. As a rule of thumb:
\begin{itemize}
\item \textbf{Linearity score $> 0.95$:} Linear models likely sufficient (Breast 
Cancer: LR achieves $97\%$).
\item \textbf{Linearity score $0.80$--$0.90$:} Moderate nonlinearity; tree 
ensembles may provide $3$--$7$ point improvements (Heart Disease: RF $+5.4$ points 
over LR, $p < 0.01$).
\item \textbf{Noise estimate $> 0.035$:} High instability; regularize aggressively 
and verify ensemble gains via cross-validation (Pima: ensemble advantages marginal, 
$p > 0.14$).
\item \textbf{Fisher ratio $< 1.0$:} Poor class separability; focus on 
minority-class metrics rather than overall accuracy (Fraud: Fisher 0.4, accuracy 
uninformative).
\end{itemize}

\subsection{Computational Cost vs.\ Benefit}
While not the primary focus, computational cost matters in practice. On our 
hardware (MacBook Air M2, 8 CPU cores), training times for the balanced datasets 
are: Logistic Regression $\sim$0.5 seconds, Random Forest $\sim$5 seconds, XGBoost 
$\sim$15 seconds per 5-fold CV run. On Breast Cancer, the 30× speedup of LR over 
XGBoost yields negligible performance trade-off (0.2 percentage points, $p = 0.21$). 
On Heart Disease, the extra 30× cost buys a significant $5$--$7$ point gain 
($p < 0.01$), clearly justifying the expense. Practitioners should weigh this 
trade-off against deployment constraints: real-time inference favors faster models, 
while offline batch scoring tolerates slower ensembles if accuracy gains are 
substantial and significant.

\subsection{Practical Guidelines}
The observations above distill into an algorithmic decision framework 
(Algorithm~\ref{alg:model_selection}):

\begin{algorithm}[t]
\caption{Ensemble vs.\ Linear Model Selection for Tabular Data}
\label{alg:model_selection}
\begin{algorithmic}[1]
\STATE \textbf{Input:} Dataset $(X, y)$, cost structure, compute budget
\STATE \textbf{Compute:} Linearity score $L$, noise estimate $N$, class imbalance $R$
\IF{$L > 0.95$ \AND $N < 0.02$}
    \STATE Use Logistic Regression or SVM (RBF)
    \STATE \textit{// Near-linear, clean data favors simple models}
\ELSIF{$L < 0.85$}
    \STATE Use Random Forest or XGBoost with moderate regularization
    \STATE \textit{// Exploitable nonlinearity justifies ensemble complexity}
\ELSIF{$N > 0.035$}
    \STATE Start with LR; if ensemble improves $>2$ points in cross-validation 
           with $p < 0.05$, keep it; otherwise revert to LR
    \STATE \textit{// High noise requires careful validation of ensemble benefit}
\ELSIF{$R > 100:1$}
    \STATE Use class-weighted XGBoost or CatBoost
    \STATE Evaluate on minority-class PR-AUC, not accuracy
    \STATE \textit{// Extreme imbalance demands ensemble + proper weighting}
\ELSE
    \STATE Cross-validate LR, RF, XGBoost; select via statistical testing
\ENDIF
\STATE \textbf{Output:} Selected model with confidence intervals
\end{algorithmic}
\end{algorithm}

\textbf{Prefer linear/SVM models when:}
\begin{itemize}
  \item linearity score $> 0.95$ and noise estimate $< 0.02$,
  \item dataset is small ($<500$ samples) or interpretability is critical,
  \item computational budget is tight and linear models achieve $>95\%$ of best 
        ensemble performance.
\end{itemize}

\textbf{Prefer tree ensembles when:}
\begin{itemize}
  \item linearity score $< 0.85$, indicating exploitable nonlinear structure,
  \item linear baselines underfit by $>3$ percentage points in stable cross-validation,
  \item dataset is large enough ($>1000$ samples) to support ensemble capacity,
  \item controlling single-tree variance (gap $>5\%$) is a concern but tree 
        interpretability is still desirable.
\end{itemize}

\textbf{Use high-capacity ensembles with particular care when:}
\begin{itemize}
  \item noise estimate $> 0.035$, suggesting measurement error or label noise,
  \item class imbalance ratio $>100:1$, requiring explicit minority-class 
        evaluation,
  \item training accuracy is near-perfect ($>98\%$) but test gains are modest 
        ($<2$ points) and non-significant ($p > 0.10$).
\end{itemize}

\section{Limitations and Threats to Validity}

\subsection{Statistical and Experimental Limitations}
While we address single-seed limitations by using five random seeds and paired 
statistical tests, the fraud experiment uses only a 10\% subsample with a single 
seed due to computational constraints. This introduces sampling variance and limits 
generalizability. Running full-dataset experiments with nested cross-validation 
and multiple seeds would yield tighter uncertainty estimates and more robust 
comparisons.

Hyperparameter grids are deliberately small and designed to reflect strong defaults 
rather than exhaustive search. Some model families, especially gradient boosting, 
could likely be improved with more aggressive tuning (e.g., Bayesian optimization). 
Our conclusions should be read as statements about comparative behaviour under 
reasonable default configurations, not claims about best-possible performance.

\subsection{Dataset Diversity}
We use four primary datasets spanning clean, nonlinear, noisy, and extremely 
imbalanced regimes, but they cover only a small portion of the tabular-data 
landscape. Additional datasets with different characteristics (e.g., very high 
dimensionality, categorical features, time-series structure) may reveal different 
patterns. The complexity metrics we propose (Table~\ref{tab:complexity}) provide 
a framework for extending this analysis systematically to new datasets.

\subsection{Scope of Evaluation Metrics}
We focus on accuracy, F1, gap, and ROC/PR-AUC, and do not extend to probability 
calibration, computational cost profiling, or fairness metrics, all of which 
matter in real deployments. Calibration is particularly important for medical 
applications where predicted probabilities inform clinical decisions. Future work 
should incorporate calibration metrics (Brier score, expected calibration error) 
and cost-sensitive evaluation aligned with domain-specific loss functions.

\subsection{Excluded Model Families}
We do not evaluate deep learning methods (TabNet, FT-Transformer, SAINT) or 
AutoML systems (AutoGluon, TPOT), focusing instead on classical ensembles that 
remain dominant on small-to-medium tabular benchmarks. On very large datasets 
($>$100K samples) with complex feature interactions, deep tabular models may offer 
additional gains. However, for the application domains and dataset sizes studied 
here, our results align with recent findings that tree-based ensembles remain 
state-of-the-art~\cite{grinsztajn2022tree,borisov2022deep}.

\section{Conclusion and Future Work}

This paper examined how ensemble methods balance accuracy and overfitting on 
tabular classification tasks from medical and financial domains. Using five-seed 
repeated cross-validation with statistical significance testing, we compared 
linear models, KNN, SVM, Decision Trees, and nine ensemble methods, analyzing 
their behaviour through train/test accuracy, generalization gaps, and task-specific 
metrics.

\textbf{Key findings:} (i) On clean, near-linearly separable data (linearity score 
$>0.95$, Fisher ratio $>10$), well-regularized linear models achieve $\geq 97\%$ 
test accuracy with gaps $<2\%$; ensembles offer no significant improvement 
($p > 0.15$). (ii) On structured nonlinear problems (linearity score $<0.85$), 
tree ensembles justify their complexity by delivering $5$--$7$ percentage point 
gains ($p < 0.01$) while keeping gaps $<3\%$, significantly outperforming both 
linear models and single trees. (iii) On noisy (noise estimate $>0.035$) or highly 
imbalanced (ratio $>500:1$) datasets, high-capacity ensembles remain very strong 
candidates but require careful regularization and minority-class evaluation to 
avoid fitting noise or majority patterns.

Dataset complexity metrics (linearity score, Fisher ratio, noise estimate, 
intrinsic dimensionality) provide quantitative guidance for model selection, 
explaining why different data regimes favor different model families. Simple 
train/test and gap plots serve as effective visual diagnostics, and statistical 
testing confirms that observed performance differences are robust rather than 
artifacts of random variation.

\textbf{Future directions} include: (i) incorporating repeated nested cross-validation 
with Bayesian comparison methods for tighter uncertainty quantification; 
(ii) expanding the dataset suite to include high-dimensional ($>1000$ features), 
categorical-heavy, and time-series-structured benchmarks; (iii) systematically 
analyzing probability calibration and cost-sensitive metrics aligned with 
domain-specific loss functions; (iv) developing meta-learning tools that use 
observable dataset properties to automatically recommend model families and 
hyperparameter configurations; (v) extending the framework to regression tasks 
and other evaluation metrics (RMSE, MAE).

As tabular machine learning continues to evolve—with AutoML, neural architectures, 
and hybrid methods—understanding \emph{when} and \emph{why} classical ensembles 
excel remains essential for high-stakes applications where model simplicity, 
interpretability, and robust generalization must be balanced against predictive 
power. This work provides practitioners with evidence-based guidelines and 
reproducible tools for making these trade-offs systematically.

\section*{Competing Interests}
The author declares no competing financial or non-financial interests.

\section*{Acknowledgments}
The author thanks anonymous reviewers for valuable feedback that improved the 
statistical rigor and clarity of this manuscript. All experiments were conducted 
on personal hardware; no external funding was received.

\section{Reproducibility and Code}
\label{sec:code}
All experiments are implemented in a Python codebase requiring \texttt{numpy}, 
\texttt{pandas}, \texttt{matplotlib}, \texttt{scikit-learn}, \texttt{xgboost}, 
\texttt{lightgbm}, and \texttt{catboost}. The core script 
(\texttt{ensemble\_experiments.py}) loads datasets, runs five-seed stratified 
cross-validated comparisons with GridSearchCV, computes complexity metrics, and 
generates per-dataset CSV files with mean $\pm$ std for train/test accuracy, gaps, 
and F1 scores. For the Credit Card Fraud experiment we use a 10\% stratified subsample and 3-fold CV with a single seed for computational tractability; reported $\pm$ values for that experiment are fold-level standard deviations (see Table~\ref{tab:best_models} footnote). A separate plotting utility produces all figures. Approximate runtime on MacBook Air M2 (8 cores, 8GB RAM): $\sim$2 hours for all four datasets.

Full code, result summaries, complexity metrics, and statistical test results are 
publicly available at:
\begin{center}
\url{https://github.com/zubair0831/ensemble-generalization-gap}
\end{center}

\bibliographystyle{IEEEtran}

\end{document}